\documentclass[9pt]{article}

\usepackage[paper=a4paper,dvips,top=3cm,left=2.5cm,right=2cm,
    foot=1cm,bottom=4cm]{geometry}

\usepackage{amsmath}
\usepackage{amssymb}
\usepackage{graphicx}   
\usepackage{verbatim}   
\usepackage{color}      
\usepackage{subfigure}  
\usepackage{stmaryrd}
\usepackage{bbm}
\usepackage{cite}
\usepackage{algorithm}
\usepackage{algorithmic}

\def\Pcal{{\mathcal P}}

\def\Hcal{{\mathcal H}}

\def\Mcal{{\mathcal M}}

\def\Vcal{{\mathcal V}}

\def\Zcal{{\mathcal Z}}

\def\kl{\mathrm{KL}}

\newcommand{\BEAS}{\begin{eqnarray*}}
\newcommand{\EEAS}{\end{eqnarray*}}
\newcommand{\BEA}{\begin{eqnarray}}
\newcommand{\EEA}{\end{eqnarray}}
\newcommand{\BEQ}{\begin{equation}}
\newcommand{\EEQ}{\end{equation}}
\newcommand{\BIT}{\begin{itemize}}
\newcommand{\EIT}{\end{itemize}}
\newcommand{\BNUM}{\begin{enumerate}}
\newcommand{\ENUM}{\end{enumerate}}

\newcommand{\EE}{\mathbb{E}}

\newcommand{\PP}{\mathbb{P}}

\newcommand{\bp}{\noindent{\textbf{Proof.}}\ }
\newcommand{\ep}{\hfill $\Box$}

\newcommand{\el}{\end{flushleft}}
\newcommand{\bl}{\begin{flushleft}}

\newcommand{\separator}{
  \begin{center}
    \rule{\columnwidth}{0.3mm}
  \end{center}
}

\newtheorem{proposition}{Proposition}
\newtheorem{theorem}{Theorem}
\newtheorem{lemma}[theorem]{Lemma}

\usepackage{amssymb}
\usepackage{amsmath}
\usepackage{graphicx}   
\usepackage{verbatim}   
\usepackage{color}      
\usepackage{subfigure}  
\usepackage{stmaryrd}
\usepackage{bbm}

\begin{document}

\title{Spectrum Bandit Optimization\footnote{A preliminary version of this work has been presented in ITW 2013 and appears in \cite{lelarge_ITW2013}.}}

\author{
			Marc Lelarge\\
			INRIA -- Ecole Normale Superieure\\
	      \textit{email: marc.lelarge@ens.fr}\\
			\\
			Alexandre Proutiere\\
			KTH Royal Institute of Technology\\
			\textit{email: alexandre.proutiere@ee.kth.se}\\
            \\
            M. Sadegh Talebi\\
            KTH Royal Institute of Technology\\
			\textit{email: mstms@kth.se}
}

\date{}

\maketitle
\begin{abstract}
We consider the problem of allocating radio channels to links in a wireless network. Links interact through interference, modelled as a conflict graph (i.e., two interfering links cannot be simultaneously active on the same channel). We aim at identifying the channel allocation maximizing the total network throughput over a finite time horizon. Should we know the average radio conditions on each channel and on each link, an optimal allocation would be obtained by solving an Integer Linear Program (ILP). When radio conditions are unknown a priori, we look for a sequential channel allocation policy that converges to the optimal allocation while minimizing on the way the throughput loss or {\it regret} due to the need for exploring sub-optimal allocations. We formulate this problem as a generic linear bandit problem, and analyze it first in a stochastic setting where radio conditions are driven by a stationary stochastic process, and then in an adversarial setting where radio conditions can evolve arbitrarily. We provide new algorithms in both settings and derive upper bounds on their regrets.
\end{abstract}




\section{Introduction}

Spectrum is a key and scarce resource in wireless communication systems, and it remains tightly controlled by regulation authorities. Most of the frequency bands are exclusively allocated to a single system licensed to use it everywhere and for periods of time that usually cover one or two decades. The consensus on this rigid spectrum management model is that it leads to significant inefficiencies in spectrum use. The explosion of demand for broadband wireless services also calls for more flexible models where much larger spectrum parts could be dynamically shared among users in a fluid manner. In such models, Dynamic Spectrum Access (DSA) techniques will play a major role. These techniques make it possible for radio devices to become frequency-agile, i.e. able to rapidly and dynamically access bands of a wide spectrum part.

In this paper, we consider wireless networks where transmitters can share a potentially large number of frequency bands or channels for transmission. In such networks, transmitters should be able to select a channel (i) that is not selected by neighbouring transmitters to avoid interference, and (ii) that offers good radio conditions. A spectrum allocation is defined by the  channels assigned to the various transmitters or links, and our fundamental objective is to devise an optimal allocation, i.e., maximizing the network-wide throughput. If the radio conditions on each link and on each channel were known, the problem would reduce to a combinatorial optimization problem, and more precisely to an Integer Linear Program. For example, if all links interfere each other (no two links can be active on the same channel), a case referred to as {\it full interference}, the optimal spectrum allocation problem is an instance of a Maximum Weighted Matching in a bipartite graph (vertices on one side correspond to links and vertices on the other side to channels; the weight of an edge, i.e., a (link, channel) pair, represents the radio conditions for the corresponding link and channel). In practice, the radio conditions on the various channels are not known a priori, and they evolve over time in an unpredictable manner. Hence, we need to dynamically learn and track the optimal spectrum allocation. This task is further complicated by the fact that we can gather information about the radio conditions for a particular (link, channel) pair only by actually including this pair in the selected spectrum allocation. We face a classical exploration vs. exploitation trade-off problem: we need to exploit the spectrum allocation with highest throughput observed so far whilst constantly exploring whether this allocation changes over time. We model our sequential spectrum allocation problem as a linear multi-armed bandit problem. The challenge in this problem resides in the very high dimension of the decision action space, i.e., in its combinatorial structure: the size of the set of possible allocations exponentially grows with the number of links and channels.

We study generic linear bandit problems in two different settings, and apply our results to sequential spectrum allocation problems. In the stochastic setting, we assume that the radio conditions for each (link,channel) pair evolve over time according to a stationary (actually i.i.d.) process whose mean is unknown. This first model is instrumental to represent scenarios where the average radio conditions evolve relatively slowly, in the sense that the spectrum allocation can be updated many times before this average exhibits significant changes. In the adversarial setting, the radio conditions evolve arbitrarily, as if they were generated by an {\it adversary}. This model is relevant when the channel allocation cannot be updated at the same pace as radio conditions change. In both settings, as usual for bandit optimization problems, we measure the performance of a given sequential decision policy through the notion of {\it regret}, defined as the difference of the performance obtained over some finite time horizon under the best static policy (i.e., assuming here that average radio conditions are known) and under the given policy. We make the following contributions:
\begin{itemize}
\item For stochastic linear bandit problems:\\
(a) we derive an asymptotic lower bound for the regret of any sequential decision policy, and show how this bound scales with number of links and channels.\\

(b) We propose a sequential decision policy for linear bandit problems, that is simple extension of the classical $\epsilon$-greedy algorithm, and provide upper bound on its regret. 

\item For adversarial linear bandit problems: We propose \textsc{ColorBand}, a new sequential decision policy, and derive an upper bound on its regret. 
\end{itemize}

{\bf Related work.} Spectrum allocation has attracted considerable attention recently, mainly due to the increasing popularity of cognitive radio systems. In such systems, transmitters have to explore spectrum to find frequency bands free from primary users. This problem can also be formulated as a bandit problem, see e.g. \cite{lai2011,ana2010}, but is simpler than our problem (in cognitive radio systems, there are basically $c$ unknown variables, each representing the probability that a channel is free). Spectrum sharing problems similar to ours have been very recently investigated in \cite{radunovic2011,gai2012combinatorial_TON}. Both aforementioned papers restrict their analysis to the case of full interference, and even in this scenario, we obtain better regret bounds. As far as we know, adversarial bandit problems have not been considered to model spectrum allocation issues. \\
There is a vast literature on bandit problems, both in the stochastic and adversarial settings, see \cite{bubek} for a quick survey. Surprisingly, there are very little work on linear bandit with discrete action space in the stochastic setting, and existing results are derived for very simple problems only, see e.g. \cite{paat2010} and references therein. In contrast, the problem has received more attention in the adversarial setting \cite{awerbuch2008, gyorgy2006, helmbold2009,kale2010,cesa2012}.

\section{Models and Objectives}

\subsection{Network and interference model}

Consider a network consisting of $n$ links indexed by $i\in [n]=\{1,\ldots,n\}$. Each link can use one of the $c$ available radio channels indexed by $j\in [c]$. Interference is represented as a conflict graph $G=(V,E)$ where vertices are links, and edges $(i,i')\in E$ if links $i$ and $i'$ interfere, i.e., these links cannot be simultaneously active. A spectrum allocation is represented as a matching $M\in \{0,1\}^{n\times c}$, where $M_{ij}=1$ if and only if link-$i$ transmitter uses channel $j$. $M$ is feasible if (i) for all $i$, the corresponding transmitter uses at most one channel, i.e., $\sum_{j\in [c]}M_{ij}\in \{ 0,1\}$; (ii) two interfering links cannot be active on the same channel, i.e., for all $i,i'\in [n]$, $(i,i')\in E$ implies for all $j\in [c]$, $M_{ij}M_{i'j}=0$  \footnote{This model assumes that the interference graph is the same over the various channels. Our analysis and results can be extended to the case where one has different interference graphs depending on the channel.}. Let ${\cal M}$ be the set of matchings. For $M\in {\cal M}$, if link $i$ is active, we denote by $M(i)$ the channel allocated to this link. We also write $(i,j)\in M$ for $i\in [n]$ and $j\in [c]$, if link $i$ is active under matching $M$, and $j=M(i)$. In the following we denote by ${\cal K}=\{ {\cal K}_{\ell}, l\in [k]\}$ the set of maximal cliques of the interference graph $G$. We also introduce $K_{\ell i}\in \{0,1\}$ such that $K_{\ell i}=1$ if and only if link $i$ belongs to the maximal clique ${\cal K}_\ell$. We will pay a particular attention to the {\it full interference} case, where the conflict graph $G$ is complete. 

\subsection{Fading}

To model the way radio conditions evolve over time on the various channels, we consider a time slotted system, where the duration of a slot corresponds either to the transmission of a single packet or to that of a fixed number $m$ of packets. The channel allocation, i.e., the chosen matching, may change at the beginning of each slot. We denote by $r_{ij}(t)$ the number of packets successfully transmitted during slot $t$ when link-$i$ transmitter selects channel $j$ for transmission in this slot and in absence of interference. Depending on the ability of transmitters to switch channels, we introduce two settings.

In the {\it stochastic setting}, the number of successful packet transmissions $r_{ij}(t)$ on link $i$ and channel $j$ are independent over $i$ and $j$, and are i.i.d. across slots $t$. The average number of successful packet transmission per slot is denoted by $\mathbb{E}[r_{ij}(t)]=\theta_{ij}$, and is supposed to be unknown initially. If $m$ packets are sent per slot, $r_{ij}(t)$ is a random variable whose distribution is that of $Y_{ij}/m$ where $Y_{ij}$ has a binomial distribution ${\rm{Bin}}(m,\theta_{ij})$. When $m=1$, $r_{ij}(t)$ is a Bernoulli random variable of mean $\theta_{ij}$. The stochastic setting models scenarios where the radio channel conditions are stationary.

In the {\it adversarial setting}, $r_{ij}(t)\in [0,1]$ can be arbitrary (as if it was generated by an {\it adversary}), and unknown in advance. This setting is useful to model scenarios where the duration of a slot is comparable to or smaller than the channel coherence time. In such scenarios, we assume that the channel allocation cannot change at the same pace as the radio conditions on the various links, which is of interest in practice, when the radios cannot rapidly change channels.

In the following, we denote by $r_M(t)$ the total number of packets successfully transmitted during slot $t$ under matching $M\in {\cal M}$, i.e.,
$$
r_M(t) = \sum_{i\in [n]} \sum_{j\in [c]} M_{ij}r_{ij}(t) =
M\bullet r(t).
$$

\subsection{Channel allocations and objectives}

We analyze the performance of adaptive spectrum allocation policies that may select different matchings at the beginning of each slot, depending on the observed received throughput under the various matchings used in the past. More precisely, at the beginning of each slot $t$, under policy $\pi$, a matching $M^\pi(t)\in {\cal M}$ is selected. This selection is made based on some feedback on the previously selected matchings and their observed throughput. We consider two types of feedback.

Under {\it semi-bandit feedback}, at the end of slot $t$, the number of packets successfully transmitted on the various links are observed, i.e., the feedback $f(t)$ is $(r_{ij}(t),i,j : M_{ij}^\pi(t)=1)$. Under {\it bandit feedback}, at the end of slot $t$, the total number of successfully sent packets is known, and so the feedback $f(t)$ is simply $r_{M^\pi(t)}(t)$. Bandit feedback is of interest when we are not able to maintain the achieved throughput per link.

At the beginning of slot $t$, the selected matching $M(t)$ may depend on past decisions and the received feedback, i.e., on $M^\pi(1),f(1),\ldots,M^\pi(t-1),f(t-1)$. The chosen matching can also be randomized (at the beginning of a slot, we sample a matching from a given distribution that depends on past observations). We denote by $\Pi$ the set of feasible policies. The objective is to identify a policy maximizing over a finite time horizon $T$ the expected number of packets successfully transmitted or simply what we call the {\it reward}. The expectation is here taken with respect to the possible randomness in the stochastic rewards (in the stochastic setting) and in the probabilistic successively selected channel allocations. Equivalently, we aim at designing a sequential channel allocation policy that minimizes the {\it regret}. The regret of policy $\pi\in \Pi$ is defined by comparing the performance achieved under $\pi$ to that of an idealised policy that assumes that the average conditions on the various links and channels are known:
\begin{equation}\label{eq:regret}
R^{\pi}(T) = \max_{M\in {\cal M}} \mathbb{E}[\sum_{t=1}^T r_{M}(t)] - \mathbb{E}[\sum_{t=1}^T r_{M^\pi(t)}(t)],
\end{equation}
where $M^\pi(t)$ denotes the matching selected in step $t$. The notion of regret quantifies the performance loss due to the need for learning radio channel conditions, and the above problem can be seen as a linear bandit problem.

\section{Optimal Static Allocation}

When evaluating the regret of a sequential spectrum allocation policy, the performance of the latter is compared to that of the best static allocation:
$$
M^\star\in\arg\max_{M\in {\cal M}} \mathbb{E}[\sum_{t=1}^T r_{M}(t)],
$$
where in the above formula, the expectation is taken with respect to the possible randomness in the throughput $r_M(t)$ (in the stochastic setting only). To simplify the presentation, we assume that the optimal static allocation $M^\star$ is unique (the analysis can be readily extended to the case where several matchings are optimal, but at the expense of the use of more involved notations). To identify $M^\star$, we have to solve an Integer Linear Program (ILP). Let us first introduce the following set of ILPs parameterized by vector $r=(r_{ij},i\in [n], j\in [c])$.
\begin{eqnarray}
\label{ILP}
\max&& \sum_{i\in [n],j\in [c]}r_{ij}M_{ij}\\
\nonumber\mbox{s.t.}&& \sum_{j\in [c]} M_{ij}\leq 1,\quad \forall i\in [n],\\
\nonumber&& \sum_{i\in [n]} K_{\ell i}M_{ij}\leq 1,\quad \forall \ell\in [k], j\in [c],\\
\nonumber&& M_{ij}\in \{0,1\},\quad \forall i\in [n], j\in [c],
\end{eqnarray}
and denote by $V(r)$ its solution. In the stochastic setting, the performance of the best static policy is then  $\mu^\star = V(\theta)$, $\theta=(\theta_{ij},i\in[n],j\in[c])$, whereas in the adversarial setting, the best static policy yields a reward equal to $V(\sum_{t=1}^T r(t))$.

\begin{lemma} The ILP problem (\ref{ILP}) is NP-complete for general interference graphs.
\end{lemma}

Indeed our ILP problem is a coloring problem of the interference graph $G$. If one considers all the links allocated to a given channel, we obtain a stable set of $G$. To be more precise, already with only one channel, our problem is NP-complete as when $c=1$ and $r_{i1}=1$ for all $i\in [n]$, then the optimum value of (\ref{ILP}) is the stable set number of the interference graph $G$ which is a NP-complete problem (Theorem 64.1 in \cite{sch04}). It should be noticed that in contrast, when the interference graph is complete, i.e., in the full interference case, the ILP problem can be interpreted as a maximum weighted matching in a bipartite graph. As a consequence, it can be solved in polynomial time \cite{sch04}.

\section{Stochastic Bandit Problem}

This section is devoted to the analysis of our linear bandit problem in the stochastic setting. We first derive an asymptotic lower bound on the regret achieved by any feasible sequential spectrum allocation policy. This provides a fundamental performance limit that no policy can beat. We then present a policy that naturally extends a classical multi-armed bandit problems to linear bandit problems, and provide upper bound on its regret. Most of the results presented here concern scenarios where semi-bandit feedback is available.

\subsection{Semi-bandit feedback}

\subsubsection{Asymptotic regret lower bound}

In their seminal paper \cite{lai1985}, Lai and Robbins consider the classical multi-armed bandit problem, where a decision maker has to sequentially select an action from a finite set of $K$ actions whose respective rewards are independent and i.i.d. across time. For example, when the rewards are distributed according to Bernoulli distributions of respective means $\theta_1,\ldots,\theta_K$, they show that the regret of any online action selection policy $\pi$ satisfies the following lower bound:
$$
\lim\inf_{T\to\infty} {R^\pi(T)\over \log(T)} \ge \sum_{i=1}^K {\theta_1 - \theta_i\over \kl(\theta_i,\theta_1)},
$$
where without loss of generality $\theta_1>\theta_i$ for all $i\neq 1$, and $\kl(u,v)$ is the KL divergence number between two Bernoulli distributions of respective means $u$ and $v$, $\kl(u,v)=u\log(u/v)+(1-u)\log(1-u)/(1-v)$. The simplicity of this lower bound is due to the stochastic independence of the rewards obtained selecting different actions. In our linear bandit problem, the rewards obtained selecting different matchings are inherently correlated (as in these matchings, a link may be allocated with the same channel). Correlations significantly complicate the derivation and the expression of the lower bound on regret. To derive such a bound, we use the techniques used in \cite{graves1997} to study the adaptive control of Markov chains.

We use the following notation: $\Theta = [0,1]^{n\times c}$; $\theta=(\theta_{ij},i\in [n],j\in [c])$; $\mu^M(\lambda) = M\bullet \lambda$, for any $M\in {\cal M}$ and $\lambda\in \Theta$. Recall that $\mu^\star = \max_{M\in {\cal M}} M\bullet \theta$, and the optimal matching is $M^\star$, i.e., $\mu^\star = M^\star\bullet \theta$. We further define: $\Delta^M=\mu^\star-\mu^M(\theta)$, $\Delta_{\min}=\min_{M\neq M^\star} \Delta^M$, and $\Delta_{\max}=\max_{M} \Delta^M$.

We introduce $B(\theta)$ as the set of {\it bad} parameters, i.e., the set of $\lambda\in \Theta$ such that matching $M^\star$ provides the same reward as under parameter $\theta$, and yet $M^\star$ is not the optimal static matching:
\begin{align*}
B(\theta)=\{ \lambda\in \Theta : (\forall i,j: M_{ij}^\star=1, \; \lambda_{ij}=\theta_{ij}),\hbox{ and } \mu^\star < \max_{M\in {\cal M}}\mu^M(\lambda)\}.
\end{align*}
Then $B(\theta)=\bigcup_{M\neq M^\star}B_M(\theta)$ where
\begin{align*}
B_M(\theta)=\{ \lambda\in \Theta : (\forall i,j: M_{ij}^\star=1, \; \lambda_{ij}=\theta_{ij}),\hbox{ and } \mu^\star < \mu^M(\lambda)\}.
\end{align*}
The reward distribution for link $i$ under matching $M$ and parameter $\theta$ is denoted by $p_i(\cdot; M,\theta)$. This distribution is over the set ${\cal S}=\{ 0,1\}$ when one packet is sent per slot, or the set ${\cal S}=\{ 0,1/m,\ldots,1\}$ if $m$ packets per slot are sent. Of course when $\sum_{j\in [c]} M_{ij}=0$, we have $p_i(0;M,\theta)=1$. When $\sum_{j\in [c]} M_{ij}=1 = M_{iM(i)}$, if a single packet is sent per slot, we have, for $y_i\in \{0,1\}$,
$$
p_i(y_i; M,\theta)=\theta_{iM(i)}^{y_i}(1-\theta_{iM(i)})^{1-y_i},
$$
and if $m$ packet are sent, we have, for $y_i\in \{ 0,1/m,\ldots,1 \}$,
$$
p_i(y_i;M,\theta)={m\choose my_i}\theta_{iM(i)}^{m y_i}(1-\theta_{iM(i)})^{m-my_i}
$$
We define the KL divergence number $\kl^M(\theta,\lambda)$ under static matching $M$ as:
\begin{align*}
\kl^M(\theta,\lambda) = \sum_{i\in [n]}\sum_{y_i\in {\cal S} }\log{p_i(y_i;M,\theta)\over p_i(y_i;M,\lambda)}  p_i(y_i;M,\theta).
\end{align*}
For instance, when a single packet is sent per slot, we get:
$$
\kl^M(\theta,\lambda) = \sum_{i\in [n]}\sum_{j\in [c]}M_{ij}\kl(\theta_{ij},\lambda_{ij}).
$$
As we shall see later in this section, we can identify sequential spectrum allocations whose regret scales as $\log(T)$ when $T$ grows large. Hence we restrict our attention to so-called {\it uniformly good} policies: $\pi\in \Pi$ is uniformly good if for all $\theta\in \Theta$, if the matching $M$ is sub-optimal ($M\neq M^\star$), then the number of times $T_M(t)$ it is selected up to time $t$ satisfies: $\mathbb{E}[T_M(t)] = o(t^\gamma)$ for all $\gamma >0$. We are now ready to state the regret lower bound.

\begin{theorem}\label{th1}
For all $\theta\in \Theta$, for all uniformly good policy $\pi\in \Pi$,
\begin{equation}\label{eq:regbound}
\lim\inf_{T\to\infty} {R^\pi(T)\over \log(T)} \ge C(\theta),
\end{equation}
where $C(\theta)$ is the optimal value of the following optimization problem:
\begin{eqnarray}
\label{eq:C_theta_opt}
&& \inf_{x_M\ge 0, M\in {\cal M}} \sum_{M\neq M^\star} x_M(\mu^\star - \mu^M(\theta))\\
\label{eq:C_theta_opt_constr}
& & \hbox{s.t. }\inf_{\lambda\in B_M(\theta)} \sum_{Q\neq M^\star}x_Q\kl^Q(\theta,\lambda) \ge 1, \;\;\forall M\neq M^\star.
\end{eqnarray}
\end{theorem}

The above lower bound is unfortunately not explicit, which motivates us to study how $C(\theta)$ scales as a function of
the problem dimensions $n$ and $c$. To this end, we introduce the following definition: Given $\theta$, we say that a set $\Hcal \subset \Mcal$ has property $P(\theta)$ iff, for all $(M,M') \in {\cal H}^2$, $M \neq M'$ we have $M_{ij} M'_{ij} (1 - M^\star_{ij}) = 0$ for all $(i,j)\in [n]\times [c]$.
We may now state Theorem~\ref{thm:Ctheta_LB}.

\begin{theorem}
\label{thm:Ctheta_LB}
Let $\Hcal$ be a maximal (inclusion-wise) subset of $\Mcal$ with property $P(\theta)$. Then:
\begin{align*}
C(\theta)\ge \sum_{M\in \Hcal} \frac{\beta(\theta)}{\max_{(i,j)\in M\setminus M^\star} \kl\left(\theta_{ij},\frac{1}{|M\setminus M^\star|}\sum_{(i',j')\in M^\star\setminus M}\theta_{i'j'}\right)},
\end{align*}
where $\beta(\theta)=\min_{M\neq M^\star}\frac{\Delta^M}{|M\setminus M^\star|}$.
\end{theorem}

\subsubsection{Regret of $\epsilon$-\textsc{Greedy} algorithms}
We investigate here the performance of a variant of the $\epsilon$-\textsc{Greedy} algorithm \cite{auer2002}. The idea of this variant is to play the matching with the best empirical reward most of the time and to explore by selecting matchings from a properly selected subset.
We introduce the following notations: Denote by $\hat{r}_{ij,s}={1\over s}\sum_{t=1}^s r_{ij}(t)$ the empirical average number of packets successfully sent over link $i$ and channel $j$ if channel $j$ has been allocated $s$ times to link $i$. Moreover, let $T_{ij}(t)$ be the number of times channel $j$ has been allocated to link $i$ up to time $t$.
Define $\hat{r}(t)=(\hat{r}_{ij,T_{ij}(t)}, i\in [n], j\in [c])$.
Finally, let ${\cal A}\subset {\cal M}$ be a set of matchings that covers all possible (link,channel) pairs. The construction of such a set is easy, and for example, in the case of full interference, we can simply use a set of $\max(n,c)$ matchings. Let $A$ be the cardinality of ${\cal A}$.

$\epsilon$-\textsc{Greedy} algorithm consists in selecting the matching that has provided with the maximum reward so far with probability $1-\epsilon_t$, and a matching selected uniformly at random among the covering set ${\cal A}$ of matchings. By reducing the exploration rate $\epsilon_t$ over time, a logarithmic regret can be achieved. More precisely, we will choose $\epsilon_t = \min(1,d/t)$ for some constant $d>0$.

\begin{algorithm}[tb]
   \caption{$\epsilon$-\textsc{Greedy}}
   \label{alg:CombUCB}
\begin{algorithmic}
   \vspace{1mm}
   \FOR{$t\geq 1$}
   \STATE Let $\epsilon_t=\min(1,d/t).$ \vspace{2mm}
   \STATE Select matching $M(t)\in \arg\max_{M\in {\cal M}} M\bullet \hat{r}(t)$ with probability $1-\epsilon_t$, and a matching uniformly selected at random in ${\cal A}$ with probability $\epsilon_t$. \vspace{2mm}
   \STATE Observe the rewards and update $\hat\theta(t)$. \vspace{2mm}
   \ENDFOR
\end{algorithmic}
\end{algorithm}

%

\begin{theorem}\label{thm:greedy} There exists a choice of parameter $d >10An^2/\Delta_{\min}^2$ such that we have:
$$
R^{\epsilon-\textsc{Greedy}}(T) \le 10A{\Delta_{\max}\over \Delta_{\min}^2}n^2\log(T)+O(1)\qquad\qquad\quad \hbox{as }T\to\infty.$$
\end{theorem}

For the case of full interference, the bound on the regret can be improved to
$$
R^{\epsilon-\textsc{Greedy}}(T) \le 10A{\Delta_{\max}\over \Delta_{\min}^2}\min(n,c)^2\log(T)+O(1).
$$
Also notice that for this case, we can select ${\cal A}$ with $A=\max(n,c)$. As a result, for the full interference case with this choice of $A$, the regret scales as ${\Delta_{\max}\over\Delta_{\min}^2}nc\min(n,c)\log(T)$ when $T$ grows large. 

For the general interference graph, we can select ${\cal A}$ with $A=\max(c,b)$ where $b$ is the minimum number of channels required to obtain a feasible channel assignment (with respect to constraints of (\ref{ILP})) in which every node $i\in [n]$ is assigned a channel. It is obvious that feasible channel assignment over conflict graph $G$ is equivalent to $b$-coloring problem of graph $G$, which is to label vertices of $G$ with $b$ colors such that no two vertices sharing the same edge have the same color. As a result, we can select $b=\gamma(G)$, where $\gamma(G)$ is the chromatic number of conflict graph $G$, and finally  $A=\max(c,\gamma(G))$. This also confirms the for the full interference case, we can select $A=\max(c,n)$ since for complete graph $G$, $\gamma(G)=n$.

\subsection{Bandit feedback}

We now briefly discuss the case where bandit feedback only is available. We derive an asymptotic lower bound for regret in this scenario, but let for future work the design of sequential spectrum allocation strategies.

Introduce for all $k=0,1,\ldots,n$:
\begin{equation}\label{eq:re_agg}
p(k;M,\theta)=\sum_{D\subset M, |D|=k} \prod_{(i,j)\in D}\theta_{ij}\prod_{(i,j)\in M\setminus D}(1-\theta_{ij}),
\end{equation}
and
$$
\kl_2^M(\theta,\lambda) = \sum_{k=0}^n p(k;M,\theta)\log {p(k;M,\theta)\over p(k;M,\lambda)}.
$$

In the following theorem, we derive an asymptotic regret lower bound. This bound is different than that derived in Theorem \ref{th1}, due to the different nature of the feedback considered. Comparing the two bounds may indicate the price to pay by restricting the set of spectrum allocation policies to those based on bandit feedback only.

\begin{theorem}
\label{th2}
For all $\theta\in \Theta$, for all uniformly good policy $\pi\in \Pi$,
\begin{equation}\label{eq:regbound}
\lim\inf_{T\to\infty} {R^\pi(T)\over \log(T)} \ge C_2(\theta),
\end{equation}
where $C_2(\theta)$ is the optimal value of the following optimization problem:
\begin{eqnarray}
&& \inf_{x_M\ge 0, M\in {\cal M}} \sum_{M\neq M^\star} x_M(\mu^\star - \mu^M(\theta)),\\
& & \hbox{s.t. }\inf_{\lambda\in B_M(\theta)} \sum_{Q\neq M^\star}x_Q\kl_2^Q(\theta,\lambda) \ge 1, \;\;\forall M\neq M^\star.
\end{eqnarray}
\end{theorem}

\section{Adversarial Bandit Problem}

In this section, we study the problem in the adversarial setting.
In \cite{acbfs02}, a regret bound of $O(\sqrt{T})$ is derived in this setting, where the constant scales as the square root of the number of arms (up to logarithmic factors) and linearly with the reward of a maximal allocation.
In our case, the number of arms typically grows exponentially with $n$ even in simple cases. For example, in the full interference case, the number of possible allocations is the number of matchings in the complete bipartite graph $([n],[c])$, i.e. $\frac{n!}{(n-c)!}$ if $n\geq c$. Also, in our case since $r_{ij}(t)\in [0,1]$, the maximal reward of an allocation is of the order $\min(n,c)$.
In the sequel, using the structure of our problem, we derive an algorithm with the same dependence in time as in \cite{acbfs02} but with much lower constants.

We start with some observations about the ILP problem (\ref{ILP}):
\BEAS
\max_{M\in \Mcal} r\bullet M &=& \max_{p(M)\geq 0, \sum_{M\in \Mcal}
  p(M)=1}\sum_{M\in \Mcal}p(M) r\bullet M\\
&=& \max_{\mu \in Co(\Mcal)} r\bullet \mu,
\EEAS
where $Co(\Mcal)$ is the convex hull of the feasible allocation matrices $\Mcal$.

We identify matrices in $\mathbb{R}^{n\times c}$ with vectors in
$\mathbb{R}^{nc}$. Without loss of generality, we can always assume that $c$ is sufficiently large
(by possibly adding artificial channels with zero reward) such that for all $i\in [n]$,
$\sum_{j\in [c]}M_{ij}=1$ for all $M\in \Mcal$, i.e. all links are
allocated to a (possibly artificial) channel. Indeed, this can be done as soon as $c\geq \gamma(G)$ where $\gamma(G)$ is the chromatic number of the interference graph $G$. In other words, the bounds derived below are valid with $c$ replaced by the maximum between the number of channels and the chromatic number of the interference graph.
With this simplifying assumption, we can embed $\mathcal{M}$ in the simplex of
distributions in $\mathbb{R}^{nc}$ by scaling all the entries by $1/n$.
Let $\Pcal$ be this scaled version of $Co(\Mcal)$.

We also define the matrix in $\mathbb{R}^{n\times c}$ with coefficients
$\mu^0_{ij}=\frac{1}{n|\Mcal|}\sum_{M\in \Mcal}M_{ij}$. Clearly
$\mu^0\in \Pcal$. We define $\mu_{\min}=\min n\mu^0_{ij}\geq \frac{1}{|\Mcal|}$.
Our algorithms are inspired from \cite{helmbold2009} where full information is revealed and uses the projection onto convex sets using the KL divergence (see Chapter 3, I-projections in \cite{csishi04}).
We denote the KL divergence between distributions $q$ and $p$ in $\Pcal$ (or more generally in the simplex of distribution in $\mathbb{R}^{nc}$) by:
\BEAS
\kl(p,q)=\sum_e p(e)\log \frac{p(e)}{q(e)},
\EEAS
where $e$ ranges over the couples $(i,j)\in [n]\times[c]$ and with the usual convention where $p\log \frac{p}{q}$ is defined to be $0$ if $p=0$ and $+\infty$ if $p>q=0$.
By definition, the projection of a distribution $q$ onto a closed convex set $\Xi$ of distributions is the $p^\star\in \Xi$ such that
\BEA\label{def:proj}
\kl(p^\star,q)=\min_{p\in \Xi}\kl(p,q).
\EEA

\subsection{Semi-bandit feedback: \textsc{ColorBand-1} Algorithm}
For the case of semi-bandit feedback, we present \textsc{ColorBand-1} algorithm, which is described next.

\begin{algorithm}[tb]
   \caption{\textsc{ColorBand-1}}
   \label{alg:ColorBand-1}
\begin{algorithmic}
   \STATE {\bf Initialization:} Start with the distribution $q_0=\mu^0$ and $\eta =\sqrt{\frac{2c\log \mu^{-1}_{\min}}{T}}$. \vspace{2mm}
   \FOR{$all\ t\ge  1$}
   \STATE Select a distribution $p_{t-1}$ over $\Mcal$ such that $\sum_{M} p_{t-1}(M) M=nq_{t-1}$. \vspace{2mm}
   \STATE Select a random matching $M(t)$ with distribution $p_{t-1}$. \vspace{2mm}
   \STATE Observe the reward matrix: $r_{ij}(t)$ for all $(i,j)\in M(t)$. \vspace{2mm}
   \STATE Construct the matrix: $\tilde{r}_{ij}(t) = \frac{1-r_{ij}(t)}{nq_{t-1}(ij)}$ for all $(i,j)\in M(t)$ and all other entries are 0. \vspace{2mm}
   \STATE Update $\tilde{q}_t(ij) \propto q_{t-1}(ij) \exp\left( -\eta \tilde{r}_{ij}(t)\right)$. \vspace{2mm}
   \STATE Set $q_t$ to be the projection of $\tilde{q}_t$ onto the set $\Pcal$ using the KL divergence. \vspace{2mm}
   \ENDFOR
\end{algorithmic}
\end{algorithm}

\begin{theorem}
\label{thm:cb1}
We have
\begin{align*}
R^{\textsc{ColorBand-1}}(T)\leq n\sqrt{2cT\log \mu^{-1}_{\min} }.
\end{align*}
\end{theorem}

Recalling that $\mu_{\min}\ge 1/|\Mcal|$, \textsc{ColorBand-1} has a regret of $O(n\sqrt{cT\log|\Mcal|})$ for the general interference graph. However, when $\mu_{\min}^{-1}=O(\mathrm{poly}(c))$, its regret is $O(n\sqrt{cT\log (c)})$.
Note that in the full interference case, we have $\mu^{-1}_{\min}=\min (n,c)$.

\bp
We first prove the following result:
\begin{lemma}\label{lem:up1}
We have for any $q\in \Pcal$,
\begin{align*}
\sum_{t=1}^T q_{t-1}\bullet \tilde{r}(t)-\sum_{t=1}^T q\bullet \tilde{r}(t) \leq
\frac{\eta}{2} \sum_{t=1}^Tq_{t-1}\bullet \tilde r^2(t) +\frac{\kl(q,q_0)}{\eta},
\end{align*}
where $\tilde{r}^2(t)$ is the vector that is the coordinate-wise
square of $\tilde{r}(t)$.
\end{lemma}

\textit{Proof:}
We have
\begin{align*}
\kl(q,\tilde{q}_t) - \kl(q,q_{t-1}) &=\sum_{e} q(e) \log
\frac{q_{t-1}(e)}{\tilde{q}_t(e)}= \eta \sum_{e}q(e)\tilde{r}_e(t)+\log Z_t,
\end{align*}
with
\begin{align*}
\log Z_t &= \log \sum_{e} q_{t-1}(e)\exp\left(-\eta
  \tilde{r}_e(n)\right)\\
&\leq \log \sum_{e} q_{t-1}(e)\left( 1-\eta \tilde{r}_e(t)+\frac{\eta^2}{2}
    \tilde{r}_e^2(t) \right)\\
&\leq -\eta q_{t-1}\bullet \tilde{r}(t) +\frac{\eta^2}{2} q_{t-1}\bullet\tilde{r}^2(t),
\end{align*}
where we used $\exp(-z)\leq 1-z+z^2/2$ for $z\ge 0$
in the first inequality and $\log (1+z)\leq z$ for all $z>-1$ in the
second inequality. We note that the use of the latter inequality is allowed, i.e. \mbox{$q_{t-1}\bullet\left(-\eta\tilde{r}(t)+\frac{\eta^2}{2}\tilde{r}^2(t)\right)>-1$}, since we have
\begin{align*}
1+q_{t-1}\bullet\left(-\eta\tilde{r}(t)+\frac{\eta^2}{2}\tilde{r}^2(t)\right)\ge \sum_{e} q_{t-1}(e)\exp\left(-\eta\tilde{r}_e(t)\right)>0.
\end{align*}

Hence, we have
\begin{align*}
\kl(q,\tilde{q}_t) - \kl(q,q_{t-1})\leq \eta q\bullet \tilde{r}(t)-\eta
q_{t-1}\bullet \tilde{r}(t) +\frac{\eta^2}{2}q_{t-1}\bullet\tilde r^2(t).
\end{align*}

Generalized Pythagorean inequality (see Theorem 3.1 in \cite{csishi04}) gives
\begin{align*}
\kl(q,q_t)+\kl(q_n,\tilde{q}_t) \leq \kl(q,\tilde{q}_t).
\end{align*}
Since $\kl(q_t,\tilde{q}_t)\geq 0$, we get
\begin{align*}
\kl(q,q_t)-\kl(q,q_{t-1})
\leq \eta q\bullet \tilde{r}(t)-\eta
q_{t-1}\bullet \tilde{r}(t) +\frac{\eta^2}{2}q_{t-1}\bullet \tilde r^2(t).
\end{align*}
Finally, summing over $n$ gives
\begin{align*}
\sum_{t=1}^T \left(q_{t-1}\bullet \tilde{r}(t)-q\bullet \tilde{r}(t)\right)
\leq
\frac{\eta}{2} \sum_{t=1}^Tq_{t-1}\bullet \tilde r^2(t) +\frac{\kl(q,q_0)}{\eta}.
\end{align*}

\ep

Let $\EE_t$ be the expectation conditioned on all the randomness
chosen by the algorithm up to time $t$.
For any $q\in\Pcal$, we have
\begin{align*}
\EE_t\left[q\bullet \tilde{r}(t)\right] = \sum_{i\in[n]}\sum_{j\in[c]}q(ij)\EE_t[\tilde{r}_{ij}(t)]=\sum_{i\in[n]}\sum_{j\in[c]}q(ij)(1-r_{ij}(t))=1-q\bullet r(t),
\end{align*}
and hence $\EE_t\left[q_{t-1}\bullet \tilde{r}(t)-q\bullet \tilde{r}(t)\right]=q \bullet r(t)-q_{t-1} \bullet {r}(t)$.

Moreover, we have
\begin{align*}
\EE_t\left[ q_{t-1}\bullet \tilde r^2(t)\right] &=\sum_{i\in
  [n]}\sum_{j\in[c]}q_{t-1}(ij)\EE_t\left[\tilde r_{ij}^2(t)\right]
  = \sum_{i\in [d]} q_{t-1}(ij)\frac{(1-r_{ij}(t))^2}{n^2q_{t-1}^2(ij)} nq_{t-1}(ij)\\
&= \sum_{i\in [n]}\sum_{j\in[c]}  \frac{(1-r_{ij}(t))^2}{n}
\leq c,
\end{align*}
since $r_{ij}(t)\in[0,1]$.


Using Lemma \ref{lem:up1} and the above bound, we get with $nq^\star$ the optimal allocation, i.e. $q^\star(e)=\frac{1}{n}$ iff $M^\star_e=1$,
\begin{align*}
R^{\textsc{ColorBand-1}}(T) &= \EE\left[ \sum_{t=1}^T nq_{t-1}\bullet \tilde{r}(t)-\sum_{t=1}^T nq^{\star}\bullet \tilde{r}(t)\right]\leq \frac{\eta ncT}{2}+\frac{n\log \mu^{-1}_{\min}}{\eta},
\end{align*}
since
\begin{align*}
\kl(q^\star,q_0) = -\frac{1}{n}\sum_{e\in M^\star}\log n\mu^{0}_{e}\leq \log \mu^{-1}_{\min}.
\end{align*}
The proof is completed by setting $\eta =\sqrt{\frac{2c\log \mu^{-1}_{\min}}{T}}$.
\ep

\subsection{Bandit feedback: \textsc{ColorBand-2} Algorithm}
We now adapt our algorithm to deal with bandit feedback. For this case, we present \textsc{ColorBand-2} algorithm, which is described as Algorithm 3.  

\begin{algorithm}[tb]
   \caption{\textsc{ColorBand-2}}
   \label{alg:ColorBand-2}
\begin{algorithmic}
   \STATE {\bf Initialization:} Start with the distribution $q_0=\mu^0$. Set $\gamma=\frac{\sqrt{n\log \mu^{-1}_{\min}}}{\sqrt{n\log \mu^{-1}_{\min}}+\sqrt{C(Cn^3c+n)T}}$ and $\eta=\gamma C$, with $C=\frac{\underline{\lambda}}{n^{3/2}}$. \vspace{2mm}
   \FOR{$all\ t\ge  1$}
   \STATE Let $q'_{t-1}=(1-\gamma)q_{t-1}+\gamma\mu^0$. \vspace{2mm}
   \STATE Select a distribution $p_{t-1}$ over $\Mcal$ such that $\sum_{M} p_{t-1}(M) M=nq'_{t-1}$. \vspace{2mm}
   \STATE Select a random matching $M(t)$ with distribution $p_{t-1}$. \vspace{2mm}
   \STATE Observe a reward $Y_t=\sum_{ij}r_{ij}(t)M_{ij}(t)$. \vspace{2mm}
   \STATE Let $\Sigma_{t-1}=\EE\left[ MM^{\top}\right]$, where $M$ has law $p_{t-1}$. Set $\tilde{r}(t) = Y_t\Sigma_{t-1}^{+}M(t)$, where   $\Sigma_{t-1}^{+}$ is the pseudo-inverse of $\Sigma_{t-1}$.\vspace{2mm}
   \STATE Update $\tilde{q}_t(ij) \propto q_{t-1}(ij) \exp\left(\eta \tilde{r}_{ij}(t)\right)$.\vspace{2mm}
   \STATE Set $q_t$ to be the projection of $\tilde{q}_t$ onto the set $\Pcal$ using the KL divergence.\vspace{2mm}
   \ENDFOR
\end{algorithmic}
\end{algorithm}

\begin{theorem}
\label{thm:cb2}
Let $\underline{\lambda}$ be the smallest nonzero eigenvalue of $\EE[MM^\top]$, where $M$ is uniformly distributed over ${\cal M}$. We have
\begin{align*}
R^{\textsc{ColorBand-2}}(T) &\le 2\sqrt{n^3T\left(nc+\frac{n^{1/2}}{\underline{\lambda}}\right)\log \mu^{-1}_{\min}}                +\frac{n^{5/2}\log \mu^{-1}_{\min}}{\underline{\lambda}}.
\end{align*}
\end{theorem}

Note that for the full interference case, we have $\underline{\lambda}=\frac{1}{n-1}$ by \cite{cesa2012}[Proposition 4], and therefore, \textsc{ColorBand-2} for $c\ge n$ has a regret of $O(\sqrt{n^4cT\log (n)})$.

\bp
We first prove a simple result:
\begin{lemma}\label{lem:proj}
For all $x\in \mathbb{R}^{nc}$, we have
$\Sigma_{t-1}^+\Sigma_{t-1}x=\overline{x}$, where $\overline{x}$ is
the orthogonal projection of $x$ onto $span(\Mcal)$, the linear space spanned by
$\Mcal$.
\end{lemma}

\textit{Proof:}
Note that for all $y\in \mathbb{R}^{nc}$, if
$\Sigma_{t-1} y=0$, then we have
\begin{align}
\label{eq:0}y^{\top}\Sigma_{t-1} y= \EE\left[ y^{\top}MM^{\top}y\right]  = \EE\left[(y^{\top}M)^2
\right]= 0,
\end{align}
where $M$ has law $p_{t-1}$ such that $\sum_{M}Mp_{t-1}(M)=q'_{n-1}$ and $q'_{t-1}=(1-\gamma)q_{t-1}+\gamma\mu^0$. By definition of $\mu^0$, each $M\in \Mcal$ has a positive probability. Hence, by (\ref{eq:0}), $y^{\top}M=0$ for all $M\in \Mcal$. In particular, we see that the linear application $\Sigma_{t-1}$
restricted to $span(\Mcal)$ is invertible and is zero on $span
(\Mcal)^{\perp}$, hence we have
$\Sigma_{t-1}^+\Sigma_{t-1} x= \overline{x}$.
\ep


\begin{lemma}\label{lem:cb2}
We have for any $\eta\le \frac{\gamma\underline{\lambda}}{n^{3/2}}$ and any $q\in \Pcal$,
\begin{align*}
\sum_{t=1}^T q\bullet \tilde{r}(t)-\sum_{t=1}^T q_{t-1}\bullet \tilde{r}(t)\leq\frac{\eta}{2} \sum_{t=1}^Tq_{t-1}\bullet \tilde r^2(t) +\frac{\kl(q,q_0)}{\eta},
\end{align*}
where $\tilde{r}^2(t)$ is the vector that is the coordinate-wise
square of $\tilde{r}(t)$.
\end{lemma}

\textit{Proof:}
We have
\begin{align*}
\kl(q,\tilde{q}_t) - \kl(q,q_{t-1}) =\sum_{e} q(e) \log
\frac{q_{t-1}(e)}{\tilde{q}_t(e)}
= -\eta \sum_{e}q(e)\tilde{r}_e(t)+\log Z_t,
\end{align*}
with
\begin{align}
\log Z_t &= \log \sum_{e} q_{t-1}(e)\exp\left(\eta
  \tilde{r}_e(t)\right)\nonumber\\
\label{eq:Zt_1}
&\leq \log \sum_{e} q_{t-1}(e)\left( 1+\eta \tilde{r}_e(t)+\eta^2
    \tilde{r}_e^2(t) \right)\\
\label{eq:Zt_2}
&\leq \eta q_{t-1}\bullet \tilde{r}(t) + \eta^2 q_{t-1}\bullet \tilde{r}^2(t),
\end{align}
where we used $\exp(z)\leq 1+ z+z^2$ for all $|z| \le 1$
in (\ref{eq:Zt_1}) and $\log (1+z)\leq z$ for all $z>-1$ in (\ref{eq:Zt_2}). We will verify later
that the choice of $\eta$ ensures that $\eta|\tilde{r}_e(t)|\le 1$ for all $e\in[n]\times [c]$.

Hence we have
\begin{align*}
\kl(q, \tilde{q}_t) - \kl(q, q_{t-1})\leq\eta q_{t-1}\bullet \tilde{r}(t)-\eta
q\bullet \tilde{r}(t) +\eta^2q_{t-1}\bullet\tilde r^2(t).
\end{align*}

Generalized Pythagorean inequality (see Theorem 3.1 in \cite{csishi04}) gives
\begin{align*}
\kl(q,q_t)+\kl(q_t,\tilde{q}_t) \leq \kl(q,\tilde{q}_t).
\end{align*}
Since $\kl(q_t,\tilde{q}_t)\geq 0$, we get
\begin{align*}
\kl(q,q_t)-\kl(q,q_{t-1}) \leq
\eta q_{t-1}\bullet \tilde{r}(t) -\eta q\bullet \tilde{r}(t) +\eta^2q_{t-1}\bullet \tilde r^2(t).
\end{align*}
Finally, summing over $t$ gives
\begin{align*}
\sum_{t=1}^T \left(q\bullet \tilde{r}(t)-q_{t-1}\bullet \tilde{r}(t)\right)\leq
\eta \sum_{t=1}^Tq_{t-1}\bullet \tilde r^2(t) +\frac{\kl(q,q_0)}{\eta}.
\end{align*}

To satisfy the condition for the inequality (\ref{eq:Zt_1}), i.e., $\eta |\tilde r_e(t)|\le 1,\; \forall e\in[n]\times [c]$, we find the upper bound for $\max_{e\in [n]\times [c]}  |\tilde{r}_e(t)|$ as follows:
\begin{align*}
\max_{e\in [n]\times [c]}|\tilde{r}_e(t)| &\le \|\tilde{r}(t)\|_{2} \nonumber\\
            &= \|\Sigma_{t-1}^+ M(t)Y_t\|_{2} \nonumber\\
            &\le n\|\Sigma_{t-1}^+ M(t)\|_{2} \nonumber\\
            &\le n\sqrt{M(t)^\top\Sigma_{t-1}^+\Sigma_{t-1}^+ M(t)} \nonumber\\
            &\le n \|M(t)\|_2\sqrt{\lambda_{\max}\left(\Sigma_{t-1}^+\Sigma_{t-1}^+\right)} \nonumber\\
				&= n^{3/2} \sqrt{\lambda_{\max}\left(\Sigma_{t-1}^+\Sigma_{t-1}^+\right)} \nonumber\\
				&= n^{3/2} \;\lambda_{\max}\left(\Sigma_{t-1}^+\right) \nonumber\\
            &= \frac{n^{3/2}}{\lambda_{\min}\left(\Sigma_{t-1}\right)},\nonumber
\end{align*}
where $\lambda_{\max}(A)$ and $\lambda_{\min}(A)$ respectively denote the maximum and the minimum nonzero eigenvalue of matrix $A$.
Note that $\mu^0$ induces uniform distribution over $\Mcal$. Thus by $q'_{t-1}=(1-\gamma)q_{t-1}+\gamma \mu^0$ we see that $p_{t-1}$ is a mixture of uniform distribution and the distribution induced by $q_{t-1}$.
Note that, we have:
\begin{align*}
\lambda_{\min}\left(\Sigma_{t-1}\right) &= \min_{\|x\|_2=1, x\in span(\Mcal)} x^{\top} \Sigma_{t-1}x.
\end{align*}
Moreover, we have
\begin{align*}
x^{\top} \Sigma_{t-1}x &= \EE\left[ x^{\top} M(t)M(t)^{\top} x\right] = \EE\left[(M(t)^{\top} x)^2 \right]\geq \gamma \EE\left[(M^{\top} x)^2 \right],
\end{align*}
where in the last inequality $M$ has law $\mu^0$. By definition, we have for any $x\in span(\Mcal)$ with $\|x\|_2=1$,
\begin{align*}
\EE\left[(M^{\top} x)^2 \right] \geq \underline{\lambda},
\end{align*}
so that in the end, we get $\lambda_{\min}(\Sigma_{t-1})\ge \gamma\underline{\lambda}$, and hence $\eta|\tilde{r}_e(t)|\le\frac{\eta n^{3/2}}{\gamma\underline{\lambda}},\; \forall e\in[n]\times [c]$. Finally, we choose $\eta\le \frac{\gamma\underline{\lambda}}{n^{3/2}}$ to satisfy the condition for the inequality we used in (\ref{eq:Zt_1}).

\ep

We have
\begin{align*}
\EE_t\left[ \tilde{r}(t)\right]= \EE_t\left[
  Y_t\Sigma_{t-1}^{+}M(t)\right]
= \EE_t\left[ \Sigma_{t-1}^{+}M(t)M(t)^{\top} r(t)\right]
= \Sigma_{t-1}^{+}\Sigma_{t-1}r(t)=\overline{r(t)},
\end{align*}
where the last equality follows from Lemma \ref{lem:proj} and
$\overline{r(t)}$ is the orthogonal projection of $r(t)$ onto
$span(\Mcal)$. In particular, for any $n'\in Co(\Mcal)$, we have
\begin{align*}
\EE_t \left[nq'\bullet \tilde{r}(t)\right] = nq'\bullet\overline{r(t)}=nq'\bullet{r(t)}.
\end{align*}

Moreover, we have:
\begin{align*}
\EE_t\left[ q_{t-1}\bullet \tilde{r}^2(t)\right]&= \sum_{e} q_{t-1}(e) \EE_t \left[\tilde{r}_e^2(t)\right]\\
&= \sum_{e} \frac{q'_{t-1}(e)-\gamma\mu^0(e)}{1-\gamma} \EE_t \left[\tilde{r}_e^2(t)\right]\\
&\le \frac{1}{n(1-\gamma)}\sum_{e} nq'_{t-1}(e) \EE_t \left[\tilde{r}_e^2(t)\right]\\
&= \frac{1}{n(1-\gamma)} \EE_t \left[\sum_{e}\tilde{M}_e(t)\tilde{r}_e^2(t)\right],
\end{align*}
where $\tilde{M}(t)$ is a random matching with the same law as $M(t)$ and independent of $M(t)$. Note that $\tilde{M}^2_e(t)=\tilde{M}_e(t)$, so that we have
\begin{align*}
\EE_t \left[\sum_{e}\tilde{M}_e(t)\tilde{r}_e^2(t)\right] &=
\EE_t\left[r(t)^{\top} M(t) M(t)^{\top} \Sigma_{t-1}^+ \tilde{M}(t) \tilde{M}(t)^{\top} \Sigma_{t-1}^+ M(t) M(t)^{\top} r(t) \right]\\
&\le n^2\EE_t[M(t)^{\top} \Sigma_{t-1}^+ M(t)],
\end{align*}
where we used the bound $M(t)^{\top} r(t)\le n$. By Lemma 15 in \cite{cesa2012}, $\EE_t[M(t)^{\top} \Sigma_{t-1}^+ M(t)]\le nc$, so that we have:
$$
\EE_t\left[ q_{t-1}\bullet \tilde{r}^2(t)\right]\le \frac{n^2c}{1-\gamma}.
$$

Observe that
\begin{align*}
\EE_t\left[ q^{\star} \bullet \tilde{r}(t)-q_{t-1}'\bullet \tilde{r}(t)\right]
&= \EE_t\left[ q^{\star} \bullet \tilde{r}(t)-(1-\gamma)q_{t-1}\bullet \tilde{r}(t)-\gamma\mu^{0}\bullet \tilde{r}(t)\right]\\
&= \EE_t\left[ q^{\star} \bullet \tilde{r}(t)-q_{t-1}\bullet \tilde{r}(t)\right]+\gamma q_{t-1}\bullet r(t)-\gamma\mu^{0}\bullet r(t)\\
&\le \EE_t\left[ q^{\star} \bullet \tilde{r}(t)-q_{t-1}\bullet \tilde{r}(t)\right]+\gamma q_{t-1}\bullet r(t)\\
&\le \EE_t\left[ q^{\star}\bullet \tilde{r}(t)-q_{t-1}\bullet \tilde{r}(t)\right]+\gamma.
\end{align*}

Using Lemma \ref{lem:cb2} and the above bounds, we get with $nq^\star$ the optimal allocation, i.e. $q^\star(e)=\frac{1}{n}$ iff $M^\star_e=1$,
\begin{align*}
R^{\textsc{ColorBand-2}}(T) &= \EE\left[ \sum_{t=1}^T nq^{\star} \bullet \tilde{r}(t)-\sum_{t=1}^T nq_{t-1}'\bullet \tilde{r}(t)\right]\\
&\le \EE\left[ \sum_{t=1}^T nq^{\star}\bullet \tilde{r}(t)-\sum_{t=1}^T nq_{t-1}\bullet \tilde{r}(t)\right]+n\gamma T\\
&\le \frac{\eta n^3cT}{1-\gamma}+\frac{n\log \mu^{-1}_{\min}}{\eta}+n\gamma T,
\end{align*}
since
\begin{align*}
\kl(q^\star,q_0) = -\frac{1}{n}\sum_{e\in M^\star}\log n\mu^{0}_{e}\leq \log \mu^{-1}_{\min}.
\end{align*}

Choosing $\eta=\gamma C$ with $C=\frac{\underline{\lambda}}{n^{3/2}}$ gives
\begin{align*}
R^{\textsc{ColorBand-2}}(T)&\le \frac{\gamma Cn^3cT}{1-\gamma}+\frac{n\log \mu^{-1}_{\min}}{\gamma C}+n\gamma T\\
        &= \frac{Cn^3c + n -n\gamma}{1-\gamma}\gamma T+\frac{n\log \mu^{-1}_{\min}}{\gamma C}\\
        &\le \frac{(Cn^3c+n)\gamma T}{1-\gamma}+\frac{n\log \mu^{-1}_{\min}}{\gamma C}.
\end{align*}
The proof is completed by setting
$
\gamma = \frac{\sqrt{n\log \mu^{-1}_{\min}}}{\sqrt{n\log \mu^{-1}_{\min}}+\sqrt{C(Cn^3c+n)T}}.
$
\ep

\subsection{Implementation}

There is a specific case where our algorithm can be efficiently
implementable: when the convex hull $Co(\Mcal)$ can be captured by
polynomial in $n$ many constraints. Note that this cannot be ensured
unless restrictive assumptions are made on the interference graph $G$
since there are up to $3^{n/3}$ maximal cliques in a graph with $n$
vertices \cite{momo65}. There are families of graphs in which the number of cliques is polynomially bounded. These families include chordal graphs, complete graphs, triangle-free graphs, interval graphs, and planar graphs. Note however, that a limited number of cliques does not ensure a priori that $Co(\Mcal)$ can be captured by a limited number of constraints. To the best of our knowledge, this problem is open and only particular cases have been solved as for the stable set polytope (corresponding to the case $c=2$, $r_{i1}=1$ and $r_{i2}=0$ with our notation) \cite{sch04}.

We consider the case where
\BEA\label{eq:CoM}
Co(\Mcal) = Co\{\forall i, \sum_{j\in [c]} M_{ij}\leq 1,\quad \forall \ell,j, \sum_{i\in [n]} K_{\ell i}M_{ij}\leq 1\}.
\EEA
Note that in the special case where $G$ is the complete graph, we have such a representation as in this case, we have
\BEAS
Co(\Mcal)=Co\{\sum_{j\in [c]} M_{ij}\leq 1,\quad \forall i,\:\sum_{i\in [n]} M_{ij}\leq 1,\quad \forall j\}.
\EEAS
We now give an algorithm for the projection of the algorithms, i.e. the projection onto $\Pcal$. Since $\Pcal$ is a scaled version of $Co(\Mcal)$, we give an algorithm for the projection onto $Co(\Mcal)$ given by (\ref{eq:CoM}).

Set $\lambda_i(0)=\mu_j(0)=0$ for all $i,j$ and then define for $t\geq 0$,
\begin{align}
\label{eq:iter1}\forall i\in [n],\:\lambda_i(t+1) =& \log \left(\sum_jM_{ij}e^{-\mu_j(t)} \right)\\
\label{eq:iter2}\forall j\in[c],\:\mu_j(t+1) =&\max_{\ell}\log\left( \sum_i K_{i\ell}M_{ij}e^{-\lambda_i(t+1)}\right).
\end{align}
We can show that
\begin{proposition}
Let $M^\star_{ij}=\lim_{t\to \infty} M_{ij}e^{-\lambda_i(t)-\mu_j(t)}$. Then $M^\star$ is the projection of $M$ onto $Co(\Mcal)$ using the KL divergence.
\end{proposition}
Although this algorithm is shown to converge, we must stress that the step (\ref{eq:iter2}) might be expensive as the number of distinct values of $\ell$ might be exponential in $n$. Again in the case of full interference, this step is easy and our algorithm reduces to Sinkhorn's algorithm (see \cite{helmbold2009} for a discussion).

\textit{Proof:}
First note that the definition of projection can be extended to non-negative vectors thanks to (\ref{def:proj}).
More precisely, given an alphabet $A$ and a vector $q\in \mathbb{R}_+^A$, we have for any probability vector $p\in \mathbb{R}_+^A$
\BEAS
\sum_{a\in A}p(a)\log \frac{p(a)}{q(a)} &\geq& \sum_ap(a)\log \frac{\sum_ap(a)}{\sum_aq(a)}\\
&=&\log \frac{1}{\|q\|_1},
\EEAS
thanks to the log-sum inequality. Hence we see that $p^\star(a)=\frac{q(a)}{\|q\|_1}$ is the projection of $q$ onto the simplex of $\mathbb{R}_+^A$.

Now define $\mathcal{A}_i=Co\{M_{ij}, \sum_j M_{ij}\leq 1\}$ and $\mathcal{B}_{\ell j}=Co\{M_{ij}, \sum_i K_{\ell i}M_{ij}\leq 1\}$. Hence $\bigcap_i\mathcal{A}_i\bigcap\bigcap_{\ell j}\mathcal{B}_{\ell j} = Co(\Mcal)$. By the argument described above, iteration (\ref{eq:iter1}) (resp. (\ref{eq:iter2})) corresponds to the projection onto $\mathcal{A}_i$ (resp. $\bigcap_\ell\mathcal{B}_{\ell j}$) and the proposition follows from Theorem 5.1 in \cite{csishi04}.

\hfill$\blacksquare$

\section{Conclusion}

In this paper, we investigate the problem of sequential spectrum allocation in wireless networks where a potentially large number of channels are available, and whose radio conditions are initially unknown. The design of such allocations has been mapped into a generic linear multi-armed bandit problem, for which we have devised efficient online algorithms. Upper bounds for the performance of these algorithms have been derived, and they are shown be as good as those of existing algorithms, both in the stochastic setting where the radio conditions on the various channels and links are modelled as stationary processes, and in the adversarial setting where no assumptions are made regarding the evolution of channel qualities. The practical implementation of our algorithms has just been briefly discussed. In particular, proposing efficient distributed implementations of these algorithms seems quite challenging, and we are currently working towards this objective.

\bibliographystyle{abbrv}
\bibliography{RA}
\appendix

\section{Proof of Theorem \ref{th1}}
To derive regret lower bounds, we apply the techniques used by Graves and Lai \cite{graves1997} to investigate efficient adaptive decision rules in controlled Markov chains. We recall here their general framework. Consider a controlled Markov chain $(X_t)_{t\ge 0}$ on a finite state space ${\cal S}$ with a control set $U$. The transition probabilities given control $u\in U$ are parameterized by $\theta$ taking values in a compact metric space $\Theta$: the probability to move from state $x$ to state $y$ given the control $u$ and the parameter $\theta$ is $p(x,y;u,\theta)$. The parameter $\theta$ is not known. The decision maker is provided with a finite set of stationary control laws $G=\{g_1,\ldots,g_K\}$ where each control law $g_j$ is a mapping from ${\cal S}$ to $U$: when control law $g_j$ is applied in state $x$, the applied control is $u=g_j(x)$. It is assumed that if the decision maker always selects the same control law $g$ the Markov chain is then irreducible with stationary distribution $\pi_\theta^g$. Now the reward obtained when applying control $u$ in state $x$ is denoted by $r(x,u)$, so that the expected reward achieved under control law $g$ is: $\mu_\theta(g)=\sum_xr(x,g(x))\pi_\theta^g(x)$. There is an optimal control law given $\theta$ whose expected reward is denoted by $\mu_\theta^{\star}=\max_{g\in G} \mu_\theta(g)$. Now the objective of the decision maker is to sequentially choose control laws so as to maximize the expected reward up to a given time horizon $T$. As for MAB problems, the performance of a decision scheme can be quantified through the notion of regret which compares the expected reward to that obtained by always applying the optimal control law.\newline

\noindent
{\bf Proof of Theorem \ref{th1}.}
We now apply the above framework to our linear bandit problem. To simplify the presentation, we consider the case where in each slot, a single packet is transmitted. We will indicate what to modify when $m$ packets are transmitted per slot.

The parameter $\theta$ takes values in $[0,1]^{n\times c}$. The Markov chain has values in ${\cal S}=\{0,1\}^n$. When $m$ packets are transmitted per slot, ${\cal S}=\{0,1/m,2/m,\ldots,1 \}^n$. The set of controls corresponds to the set of matchings ${\cal M}$, and the set of control laws is also ${\cal M}$. These laws are constant, in the sense that the control applied by control law $M$ does not depend on the state of the Markov chain, and corresponds to selecting matching $M$. The transition probabilities are given as follows: for all $x,y\in {\cal S}$,
$$
p(x,y;M,\theta)=p(y;M,\theta)=\prod_{i\in [n]}p_i(y_i;M,\theta),
$$
where for all $i\in [n]$, if $\sum_{j\in [c]}M_{ij}=0$, $p_i(0;M,\theta)=1$, and if $\sum_{j\in [c]}M_{ij}=M_{iM(i)}=1$, $p_i(y_i;M,\theta)=\theta_{iM(i)}^{y_i}(1-\theta_{iM(i)})^{1-y_i}$. When $m$ packets are sent per slot, the last formula has to be replaced by: $p_i(y_i;M,\theta)={m\choose my_i}\theta_{iM(i)}^{m y_i}(1-\theta_{iM(i)})^{m-my_i}$. Finally, the reward $r(y,M)$ is defined by $r(y,M)=M\bullet y$. Note that the state space of the Markov chain is here finite, and so, we do not need to impose any cost associated with switching control laws (see the discussion on page 718 in \cite{graves1997}).

We can now apply Theorem 1 in \cite{graves1997}. Note that the KL number under matching $M$ is:
\begin{align*}
\kl^M(\theta,\lambda) & = \sum_y \log{p(y;M,\theta)\over p(y;M,\lambda)}  p(y;M,\theta)\\
& = \sum_{i\in [n]}\sum_{y_i\in S}\log{p_i(y_i;M,\theta)\over p_i(y_i;M,\lambda)}  p_i(y_i;M,\theta),
\end{align*}
where $S=\{ 0,1\}$ when one packet is sent per slot and $S=\{ 0,1/m,\ldots,1\}$ if $m$ packets per slot are sent.

For instance, when a single packet is sent per slot, we get:
$$
\kl^M(\theta,\lambda) = \sum_{i\in [n]}\sum_{j\in [c]}M_{ij}\kl(\theta_{ij},\lambda_{ij}),
$$
where $\kl(u,v)=u\log(u/v) +(1-u)\log((1-u)/(1-v))$. From Theorem 1 in \cite{graves1997}, we conclude that for any uniformly good rule $\pi$,
$$
\lim\inf_{T\to\infty} {R^\pi(T)\over \log(T)} \ge C(\theta),
$$
where $C(\theta)$ is the optimal value of the following optimization problem:
\begin{eqnarray}
&& \inf_{x_M\ge 0, M\in {\cal M}} \sum_{M\neq M^\star} x_M(\mu^\star - \mu^M(\theta)),\\
& & \hbox{s.t. }\inf_{\lambda\in B(\theta)} \sum_{Q\neq M^\star}x_Q\kl^Q(\theta,\lambda) \ge 1.
\end{eqnarray}
The result is obtained by observing that $B(\theta)=\bigcup_{M\neq M^\star}B_M(\theta)$.

\ep

\noindent
{\bf Proof of Theorem \ref{th2}.} In the case of bandit feedback, when matching $M$ is selected at time $t$, the global reward $r_M(t)$ only is known. To take this limited feedback into account, the state space of the corresponding Markov chain should record the aggregate reward only. Hence, we have ${\cal S}=\{0,1,\ldots,n\}$. When the state is $k$, it means that the global received reward is equal to $k$. The probability that the reward under matching $M$ is equal to $k$ is then $p(k;M,\theta)$ defined in (\ref{eq:re_agg}), and so: for all $k',k\in {\cal S}$,
$$
p(k',k;M,\theta)=p(k;M,\theta).
$$
Theorem \ref{th2} is then a direct consequence of Theorem 1 in \cite{graves1997}.\ep

\section{Proof of Theorem \ref{thm:Ctheta_LB}}
The proof proceeds in three steps. In the subsequent analysis, given the optimization problem \textsf{P}, we use $\textrm{val}(\textsf{P})$ to denote its optimal value.

\paragraph{\underline{Step 1}.}
In this step, first we introduce an equivalent formulation for problem (\ref{eq:C_theta_opt}) by simplifying its constraints. We show that constraint (\ref{eq:C_theta_opt_constr}) is equivalent to:
\begin{align*}
\inf_{\lambda\in B_M(\theta)}\sum_{\ell\in M\setminus M^\star}\kl(\theta_\ell,\lambda_\ell)\sum_{Q\neq M^\star} Q_\ell x_Q\ge 1,\;\; \forall M\neq M^\star.\nonumber
\end{align*}
Observe that:
\begin{align*}
\sum_{Q\neq M^\star} x_Q\kl^Q(\theta,\lambda)=\sum_{Q\neq M^\star} x_Q\sum_{\ell\in [n]\times [c]} Q_{\ell}\kl(\theta_{\ell},\lambda_{\ell})
=\sum_{\ell}\kl(\theta_\ell,\lambda_\ell)\sum_{Q\neq M^\star} Q_\ell x_Q.
\end{align*}
Fix $M\neq M^\star$. In view of the definition of $B_M(\theta)$, we can find $\lambda\in B_M(\theta)$ such that $\lambda_\ell=\theta_\ell, \forall \ell\in ([n]\times [c]\setminus M)\cup M^\star$. Thus, for the r.h.s. of the $M$-th constraint in (\ref{eq:C_theta_opt_constr}), we get:
\begin{align}
\inf_{\lambda\in B_M(\theta)} \sum_{Q\neq M^\star}x_Q\kl^Q(\theta,\lambda) &=
\inf_{\lambda\in B_M(\theta)} \sum_{\ell\in [n]\times [c]}\kl(\theta_\ell,\lambda_\ell)\sum_{Q\neq M^\star} Q_\ell x_Q\nonumber\\
&= \inf_{\lambda\in B_M(\theta)}\sum_{\ell \in M\setminus M^\star}\kl(\theta_\ell,\lambda_\ell)\sum_{Q} Q_\ell x_Q,\nonumber
\end{align}
and therefore problem (\ref{eq:C_theta_opt}) can be equivalently written as:
\begin{align}
\label{eq:C_theta_opt_alter}
C(\theta)=& \inf_{x_M\ge 0, M\in {\cal M}} \sum_{M\neq M^\star} x_M(\mu^\star - \mu_M(\theta)),\\
&  \hbox{s.t. }\inf_{\lambda\in B_M(\theta)}\sum_{\ell \in M\setminus M^\star}\kl(\theta_\ell,\lambda_\ell)\sum_{Q} Q_\ell x_Q \ge 1,\;\; \forall M\neq M^\star.
\end{align}

Next, we formulate an LP whose value gives a lower bound for $C(\theta)$. 
Define $\hat\lambda(M)=(\hat\lambda_\ell(M), \ell\in [n]\times [c])$ with
\[ \hat\lambda_{\ell}(M) = \left\{
  \begin{array}{l l}
    \frac{1}{|M\setminus M^\star|}\sum_{k\in M^\star\setminus M}\theta_k & \quad \hbox{if }\ell\in M\setminus M^\star\\
    \theta_\ell & \quad \textrm{otherwise.}
  \end{array} \right.\]
Clearly $\hat\lambda(M)\in B_M(\theta)$, and therefore:
\begin{align*}
\inf_{\lambda\in B_M(\theta)}\; \sum_{\ell\in M\setminus M^\star} \kl(\theta_\ell,\lambda_\ell)\sum_{Q} Q_\ell x_Q&\le \sum_{\ell\in M\setminus M^\star} \kl(\theta_\ell,\hat\lambda_\ell(M))\sum_{Q} Q_\ell x_Q.
\end{align*}

Then, we can write:
\begin{align}
\label{eq:Ctheta_opt_alter2}
C(\theta)\ge& \inf_{x\ge 0} \sum_{M\neq M^\star} \Delta^Mx_M\\
& \hbox{s.t. } \sum_{\ell\in M\setminus M^\star} \kl(\theta_\ell,\hat\lambda_\ell(M))
\sum_{Q} Q_\ell x_Q \ge 1,\;\;\; \forall M\neq M^\star.
\end{align}

For any $M\neq M^\star$ introduce: $g_M=\max_{\ell\in M\setminus M^\star} \kl(\theta_\ell,\hat\lambda_\ell(M))$. Now we form \textsf{P1} as follows:
\begin{align}
\textsf{P1:}\quad& \inf_{x\ge 0} \;\sum_{M\neq M^\star} \Delta^M x_M\\
& \hbox{s.t. } \sum_{\ell \in M\setminus M^\star} \sum_{Q} Q_\ell x_Q\geq \frac{1}{g_M},\quad \forall M\neq M^\star.
\end{align}
Observe that $C(\theta)\ge \textrm{val}(\textsf{P1})$ since the feasible set of problem (\ref{eq:Ctheta_opt_alter2}) is contained in that of \textsf{P1}.

\paragraph{\underline{Step 2.}}
In this step, we formulate an LP to give a lower bound for val(\textsf{P1}).
To this end, for any sub-optimal edge $\ell\in [n]\times [c]\setminus M^\star$, we define $z_\ell=\sum_{M} M_\ell x_M$. Further, we let  $z=[z_\ell, \ell\in[n]\times [c]\setminus M^\star]$.
Next, we represent the objective of \textsf{P1} in terms of $z$, and give a lower bound for it as follows:

\begin{align*}
\sum_{M\neq M^\star} \Delta^M x_M&= \sum_{M\neq M^\star} x_M\sum_{\ell \in M\setminus M^\star} \frac{\Delta^M}{|M\setminus M^\star|} \\
						&=\sum_{M\neq M^\star} x_M\sum_{\ell\in [n]\times [c]\setminus M^\star} \frac{\Delta^M}{|M\setminus M^\star|}M_\ell \\
			&\ge\min_{M\neq M^\star} \frac{\Delta^M}{|M\setminus M^\star|}\cdot\sum_{\ell\in [n]\times [c]\setminus M^\star} \sum_{M'\neq M^\star} M'_\ell x_{M'}  \\
			&=\min_{M\neq M^\star} \frac{\Delta^M}{|M\setminus M^\star|}\cdot\sum_{\ell\in [n]\times [c]\setminus M^\star} z_{\ell}\\
            &= \beta(\theta) \sum_{\ell\in [n]\times [c]\setminus M^\star} z_{\ell}.
\end{align*}
Then, defining
\begin{align*}
\textsf{P2:}&\quad  \inf_{z\ge 0} \beta(\theta)\sum_{\ell\in [n]\times [c]\setminus M^\star} z_\ell\\
& \hbox{s.t. } \sum_{\ell \in M\setminus M^\star} z_{\ell}\geq \frac{1}{g_M},\;\; \forall M\neq M^\star,
\end{align*}
yields:
$\textrm{val}(\textsf{P1})\ge \textrm{val}(\textsf{P2}).$

\paragraph{\underline{Step 3.}}
Introduce set $\Hcal$ satisfying property $P(\theta)$ as stated in Section 4.
Now define
$$
\Zcal=\left\{z\in \mathbb R_+^{nc-n}:
\sum_{\ell\in M\setminus M^\star} z_{\ell}\geq \frac{1}{g_M},\;\; \forall M
\in \Hcal\right\},
$$
and
\begin{align*}
\textsf{P3:}\quad \inf_{z\in\Zcal} \beta(\theta)\sum_{\ell\in [n]\times [c]\setminus M^\star} z_{\ell}.
\end{align*}
Observe that $\textrm{val}(\textsf{P2})\ge\textrm{val}(\textsf{P3})$ since the feasible set of \textsf{P2} is contained in $\Zcal$.

It can be easily seen that:
\begin{align*}
\textrm{val}(\textsf{P3})&= \sum_{M\in\Hcal} \frac{\beta(\theta)}{g_M}\\
                         &\ge \sum_{M\in\Hcal} \frac{\beta(\theta)}{\max_{\ell\in M\setminus M^\star} \kl(\theta_\ell,\hat\lambda_\ell(M))} \\
                         & =  \sum_{M\in\Hcal} \frac{\beta(\theta)}{\max_{\ell\in M\setminus M^\star} \kl\left(\theta_\ell,\frac{1}{|M\setminus M^\star|}\sum_{k\in M^\star\setminus M}\theta_k\right)}.
\end{align*}

The proof is completed by observing that:
$C(\theta)\ge \textrm{val}(\textsf{P1})\ge \textrm{val}(\textsf{P2})\ge \textrm{val}(\textsf{P3})$.
\ep

\section{Proof of Theorem \ref{thm:greedy}}

We first provide a bound on the probability of choosing a sub-optimal matching $M$. In what follows, we denote $X^M(t)=\sum_{i\in\Vcal_M} \hat{r}_{iM(i),T_{iM(i)}(t)}$, where
$$\mathcal V_M=\{i\in[n]:M_{ij}=1\quad\hbox{for some}\quad j\in [c]\}$$
with cardinality $V_M$.

For $t>d$, the probability of choosing a sub-optimal matching $M$ can be written as
$$
\PP\left[I_t=M\right]\leq \frac{\epsilon_t}{A}\mathbbmss 1\{M\in\mathcal A\}+(1-\epsilon_t)F,
$$
where $F=\PP\left[ X^{M}(t-1)\geq X^{M^\star}(t-1)\right].$ Then
\begin{eqnarray}
F&\leq& \PP\left[\sum_{i\in\mathcal V_M} \hat{r}_{iM(i),T_{iM(i)}(t-1)} \geq \sum_{i\in\mathcal V_M} \left(\theta_{iM(i)}+\frac{\Delta^M}{2V_M}\right)\right]\nonumber\\
&+&\PP\left[\sum_{i\in\mathcal V_{M^\star}} \hat{r}_{iM^\star(i),T_{iM^\star(i)}(t-1)} \leq \sum_{i\in\mathcal V_{M^\star}} \left(\theta_{iM^\star(i)}-\frac{\Delta^M}{2V_{M^\star}}\right)\right].\nonumber
\end{eqnarray}
Using the union bound we get
\begin{eqnarray}
\PP\left[\sum_{i\in\Vcal_M} \hat{r}_{iM(i),T_{iM(i)}(t-1)} \geq \sum_{i\in\Vcal_M} \left(\theta_{iM(i)}+\frac{\Delta^M}{2V_M}\right)\right]
\leq\sum_{i\in\Vcal_M} \underbrace{\PP\left[\hat{r}_{iM(i),T_{iM(i)}(t-1)} \geq \theta_{iM(i)}+\frac{\Delta^M}{2V_M}\right\}}_{\PP\{\mathcal B_i(t-1)\}}.\nonumber
\end{eqnarray}
Now for $i\in\mathcal V_M$, using Chernoff-Hoeffding bound we get
\begin{align}\label{eq:Pr_B1_IT1}
\PP[\mathcal B_i(t)]\leq  \sum_{s=1}^{t} e^{-\frac{s}{2}\left(\frac{\Delta^M}{V_M}\right)^2} \PP\left[T_{iM(i)}(t)=s : \hat{r}_{iM(i),s}\geq
\theta_{iM(i)}+\frac{\Delta^M}{2V_M}\right].
\end{align}
Observe that for $w>0$, $\sum_{s=x+1}^t e^{-ws}< \frac{1}{w}e^{-wx}$. This implies $\sum_{s=x+1}^t e^{-\frac{s}{2}\left(\frac{\Delta^M}{V_M}\right)^2} \leq 2\left(\frac{V_M}{\Delta^M}\right)^2 e^{-\frac{x}{2}\left(\frac{\Delta^M}{V_M}\right)^2}$. Let $y_0=\frac{1}{2 A}\sum_{s=1}^{t}\epsilon_s$. We have:
\begin{align*}
&\sum_{s=\lfloor y_0\rfloor+1}^{t} e^{-\frac{s}{2}\left(\frac{\Delta^M}{V_M}\right)^2} \PP\left[T_{iM(i)}(t)=s : \hat{r}_{iM(i),s}\geq \theta_{iM(i)}+\frac{\Delta^M}{2V_M}\right] \le 2\left(\frac{V_M}{\Delta^M}\right)^2 e^{-\frac{\lfloor y_0 \rfloor}{2}\left(\frac{\Delta^M}{V_M}\right)^2}.
\end{align*}
We also have:
\begin{align*}
&\sum_{s=1}^{\lfloor y_0\rfloor} e^{-\frac{s}{2}\left(\frac{\Delta^M}{V_M}\right)^2} \PP\left[T_{iM(i)}(t)=s : \hat{r}_{iM(i),s}\geq \theta_{iM(i)}+\frac{\Delta^M}{2V_M}\right]\\
&\le \sum_{s=1}^{\lfloor y_0\rfloor} \PP\left[T_{iM(i)}(t)=s : \hat{r}_{iM(i),s}\geq \theta_{iM(i)}+\frac{\Delta^M}{2V_M}\right]\\
&\le \sum_{s=1}^{\lfloor y_0\rfloor} \PP\left[T^R_{iM(i)}(t)\leq s : \hat{r}_{iM(i),s}\geq \theta_{iM(i)}+\frac{\Delta^M}{2V_M}\right]\\
& \le \sum_{s=1}^{\lfloor y_0\rfloor} \PP\left[T^R_{iM(i)}(t)\leq s\right] \\
& \le y_0 \PP\left[T_{iM(i)}^R(t)\leq y_0\right],
\end{align*}
where $T_{iM(i)}^R(t)$ denotes the number of times that pair $(i,M(i))$ is chosen during the exploration phase up to time $t$, and where we used the fact that $T_{iM(i)}^R(t)\leq T_{iM(i)}(t)$ implies \mbox{$\PP[T_{iM(i)}(t)\leq y_0]\leq \PP\left[T_{iM(i)}^R(t)\leq y_0\right]$}. Now it can be easily shown that:
$$
\mathbb E \big[T_{iM(i)}^R(t)\big]=\frac{1}{A}\sum_{s=1}^t\epsilon_s=2y_0,
$$
$$
\textrm{Var}\big[T_{iM(i)}^R(t)\big]=\sum_{s=1}^t\frac{\epsilon_s}{A}\left(1-\frac{\epsilon_s}{A}\right)\leq \frac{1}{A}\sum_{s=1}^t\epsilon_s=2y_0.
$$
Using Bernstein's inequality, we then have $\PP\left[T_{iM(i)}^R(t)\leq y_0\right]\leq e^{-\frac{y_0}{5}}$. Finally, using $\epsilon_s=\min(1,d/s)$, we get $y_0=\frac{d}{2A}+\frac{d}{2A}\log\frac{t}{d}=\frac{d}{2A}\log\frac{et}{d}$. In summary, we proved that:
\begin{align*}
\PP[\mathcal B_i(t)]\leq&
y_0 e^{-\frac{y_0}{5}}+2\left(\frac{V_M}{\Delta^M}\right)^2 e^{-\frac{\lfloor y_0 \rfloor}{2}\left(\frac{\Delta^M}{V_M}\right)^2} \nonumber\\
=&
\frac{d}{2A}\log\frac{et}{d}.e^{-\frac{d}{10A}\log\frac{et}{d}}+ 2\left(\frac{V_M}{\Delta^M}\right)^2 e^{-\frac{d}{4A} \left(\frac{\Delta^M}{V_M}\right)^2\log\frac{et}{d}}\nonumber\\
=&h(t)t^{-\frac{d}{10A}}+G_Mt^{{-\frac{d}{4A}\left(\frac{\Delta^M}{V_M}\right)^2}},
\end{align*}
where \mbox{$h(t)\triangleq\frac{d}{2A}\left(\frac{e}{d}\right)^{-\frac{d}{10A}} \log\left(\frac{et}{d}\right)$}
and \mbox{$G_M\triangleq 2\left(\frac{V_M}{\Delta^M}\right)^2 \left(\frac{e}{d}\right)^{-\frac{d}{4A}\left(\frac{\Delta^M}{V_M}\right)^2}$}. As a result, we obtain:
\begin{align*}
\PP\left[\sum_{i\in\Vcal_M} \hat{r}_{iM(i),T_{iM(i)}(t-1)} \geq \sum_{i\in\Vcal_M} \left(\theta_{iM(i)}+\frac{\Delta^M}{2V_M}\right)\right] &\leq \sum_{i\in\Vcal_M} \PP[\mathcal B_i(t-1)]\\
&\leq V_Mh(t-1)(t-1)^{-\frac{d}{10A}}+V_MG_M\cdot(t-1)^{-\frac{d}{4A}\left(\frac{\Delta^M}{V_M}\right)^2}.
\end{align*}

It can be shown similarly that
\begin{align*}
\PP\left[\sum_{i\in\mathcal V_{M^\star}} \hat{r}_{iM^\star(i),T_{iM^\star(i)}(t-1)} \leq \sum_{i\in\mathcal V_{M^\star}} \left(\theta_{iM^\star(i)}-\frac{\Delta^M}{2V_{M^\star}}\right)\right]
\leq V_{M^\star}h(t-1)(t-1)^{-\frac{d}{10A}}+V_{M^\star}G_{M^\star}(t-1)^{-\frac{d}{4A}\left(\frac{\Delta^M}{V_{M^\star}}\right)^2},
\end{align*}
where \mbox{$G_{M^\star}=2\left(\frac{V_{M^\star}}{\Delta^M}\right)^2 \left(\frac{e}{d}\right)^{-\frac{d}{4A}\left(\frac{\Delta^M}{V_{M^\star}}\right)^2}$}.

We conclude that:
\begin{align*}
\PP[I_t=M]&\leq\frac{d}{tA}\mathbbmss 1\{M\in\mathcal A\}+(V_{M^\star}+V_M)\left(1-\frac{d}{t}\right)h(t-1)(t-1)^{-\frac{d}{10A}}\nonumber\\
&+V_MG_M\left(1-\frac{d}{t}\right)(t-1)^{-\frac{d}{4A}\left(\frac{\Delta^M}{V_M}\right)^2}
+V_{M^\star}G_{M^\star}\left(1-\frac{d}{t}\right)(t-1)^{-\frac{d}{4A}\left(\frac{\Delta^M}{V_{M^\star}}\right)^2}.
\end{align*}

The upper bound on regret is then
\begin{align}
R^{\epsilon-\textsc{Greedy}}(t)=&\sum_{s=1}^{t}\sum_{M\neq M^\star} \Delta^M\PP[I_s=M]\nonumber\\
\leq&\Delta_{\max}\sum_{s=1}^t\sum_{M\neq M^\star} \PP[I_s=M]\nonumber\\
\leq& \Delta_{\max} \sum_{s=1}^t\sum_{M\neq M^\star}\frac{d}{sA}\mathbbmss 1\{M\in\mathcal A\}\nonumber\\
+&\Delta_{\max}\sum_{s=1}^t\sum_{M\neq M^\star}(V_M+V_{M^\star})\left(1-\frac{d}{s}\right)h(s-1)(s-1)^{-\frac{d}{10A}}\nonumber\\
+&\Delta_{\max}\sum_{s=1}^t\sum_{M\neq M^\star}V_MG_M\left(1-\frac{d}{s}\right)(s-1)^{-\frac{d}{4A}\left(\frac{\Delta^M}{V_M}\right)^2}\nonumber\\
\label{eq:R_eps_greedy_I1}
+&\Delta_{\max}\sum_{s=1}^t\sum_{M\neq M^\star}
V_{M^\star}G_{M^\star}\left(1-\frac{d}{s}\right)(s-1)^{-\frac{d}{4A}\left(\frac{\Delta^M}{V_{M^\star}}\right)^2}.
\end{align}

Observe that $d>10An^2/\Delta_{\min}^2$ implies $d>10A$. Note further that $V_M\leq n, \forall M$. Then $d>10An^2/\Delta_{\min}^2$ implies that for any $M\neq M^\star$:
$$
d>4A\left(\frac{V_M}{\Delta^M}\right)^2\quad\textrm{and}\quad d>4A\left(\frac{V_{M^\star}}{\Delta^M}\right)^2.
$$
As a result, in the r.h.s. of (\ref{eq:R_eps_greedy_I1}), except the first term, the others will be bounded as $t$ grows large.
Then, after simplifications, we get:
\begin{eqnarray}
R^{\epsilon-\textsc{Greedy}}(t)&\leq&  d\Delta_{\max}\log t+O(1),\quad \hbox{as }t\to\infty.
\end{eqnarray}
The proof is completed by taking the infimum in the r.h.s. over $d>10An^2/\Delta_{\min}^2$.

\ep

\end{document}